\documentclass[10pt,twocolumn,letterpaper]{article}

\usepackage[pagenumbers]{iccv} 

\usepackage{times}
\usepackage{epsfig}
\usepackage{graphicx}
\usepackage{amsmath}
\usepackage{amssymb}
\usepackage{colortbl}
\usepackage{array}
\usepackage{calc}
\usepackage{booktabs}
\usepackage{float}
\usepackage{bm}
\usepackage{multicol}
\usepackage{algorithm}
\usepackage[noend]{algpseudocode}
\usepackage{algorithmicx}
\usepackage{color}
\usepackage{multirow}
\usepackage{cuted}
\usepackage{capt-of}
\usepackage{subcaption}
\usepackage{pifont}
\usepackage[symbol]{footmisc}
\newcommand{\norm}[1]{\left\lVert#1\right\rVert}
\usepackage[english]{babel}
\usepackage{amsthm}
\usepackage{bbding}

\newtheorem{theorem}{Theorem}[section]
\newtheorem{lemma}[theorem]{Lemma}
\newtheorem{assumption}{Assumption}

\definecolor{iccvblue}{rgb}{0.21,0.49,0.74}
\usepackage[pagebackref,breaklinks,colorlinks,allcolors=iccvblue]{hyperref}

\def\paperID{4210} 
\def\confName{ICCV}
\def\confYear{2025}

\title{Optimal Stepsize for Diffusion Sampling}

\author{Jianning Pei$^{1,2}$, Han Hu$^{2}$, Shuyang Gu$^{2}\footnotemark[2]$\\
$^{1}$University Chinese Academic of Science\ \ \ $^{2}$Tencent Hunyuan Research\\
{\tt\small jianningpei22@mails.ucas.ac.cn, cientgu@tencent.com}
}

\begin{document}
\maketitle

\renewcommand{\thefootnote}{\fnsymbol{footnote}}  
\footnotetext[2]{Corresponding Author.}

\begin{strip}\centering
\vspace{-0.6cm}
\includegraphics[width=\textwidth]{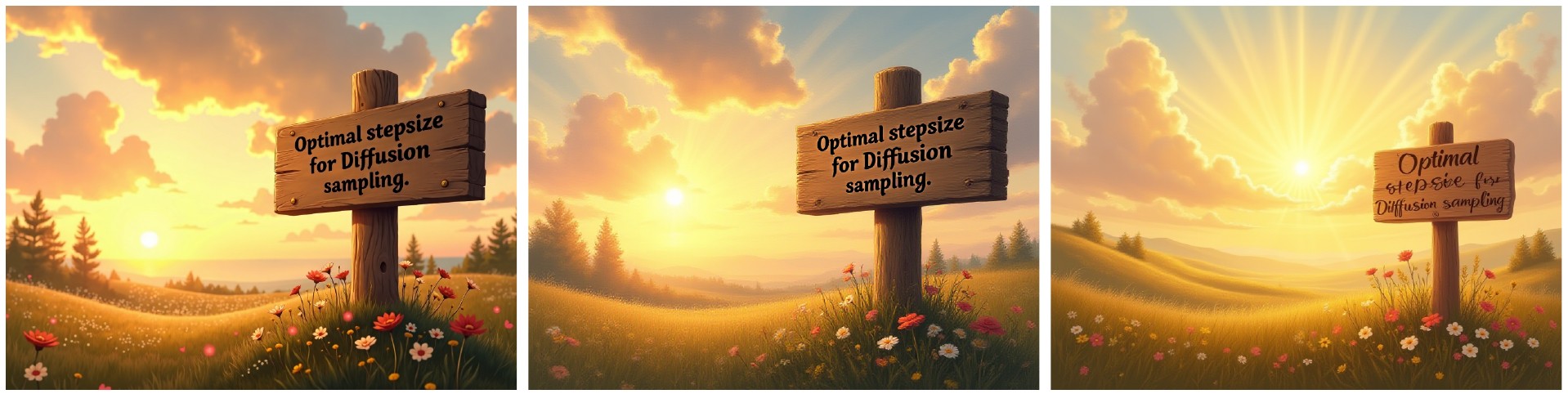}
\vspace{-0.6cm}
\captionof{figure}{Flux sampling results using different stepsize schedules. Left: Original sampling result using $100$ steps. Middle: Optimal stepsize sampling result within $10$ steps. Right: Naively reducing sampling steps to $10$.  
\label{fig:feature-graphic}}
\end{strip}

\begin{abstract}
        Diffusion models achieve remarkable generation quality but suffer from computational intensive sampling due to suboptimal step discretization. While existing works focus on optimizing denoising directions, we address the principled design of stepsize schedules. This paper proposes Optimal Stepsize Distillation, a dynamic programming framework that extracts theoretically optimal schedules by distilling knowledge from reference trajectories. By reformulating stepsize optimization as recursive error minimization, our method guarantees global discretization bounds through optimal substructure exploitation. Crucially, the distilled schedules demonstrate strong robustness across architectures, ODE solvers, and noise schedules. Experiments show \textbf{10$\times$} accelerated text-to-image generation while preserving \textbf{99.4\%} performance on GenEval. Our code is available at \url{https://github.com/bebebe666/OptimalSteps}.
\end{abstract}

\section{Introduction}
    Score-based generative models \cite{scorematching,song2020improved} have emerged as a groundbreaking paradigm in generative modeling, achieving state-of-the-art results across diverse domains. Unlike traditional approaches that directly model complex data distributions, these methods decompose the target distribution into a sequence of conditional distributions structured as a Markov chain.  While this decomposition enables tractable training through iterative noise prediction, it inherently necessitates computational expensive sampling procedures.

    From a theoretical perspective, diffusion models approximate the data generation process as a continuous transformation from noise to structured data, parameterized by an infinite series of conditional distributions mapping. But in practice, computational constraints require discretizing this continuous trajectory into a finite sequence of steps, introducing a critical trade-off between sampling efficiency and approximation fidelity.

\begin{figure}
    \vspace{-0.6cm}
        \centering
        \includegraphics[width=0.49\textwidth]{
            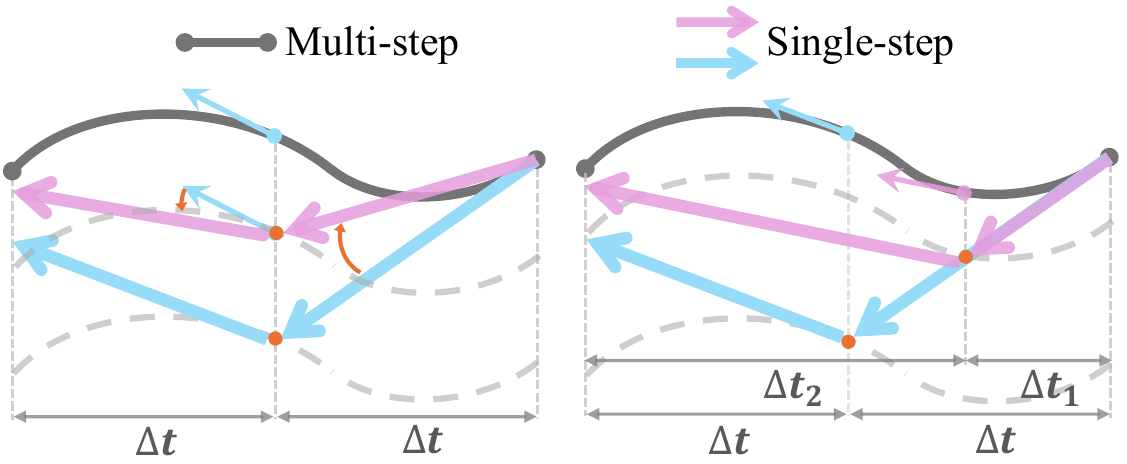
        }
        \vspace{-0.6cm}
        \caption{Two key factors in diffusion sampling: direction strategy(left) and stepsize strategy(right).}
        \label{fig:introfig}
        \vspace{-0.6cm}
    \end{figure}

    The discretization includes two key factors: the update direction (determined by the score function) and the stepsize (controlling the progression towards the target distribution), as shown in Figure \ref{fig:introfig}. While significant attention has been devoted to optimizing update directions, including higher-order solvers~\cite{dpm,dpm++} and corrected network predictions~\cite{unipc}, the principled design of stepsize remains underexplored. Existing methods predominantly rely on heuristic schedules, such as the uniform timesteps in DDIM~\cite{ddim} or the empirically designed schedule in EDM~\cite{edm}, lacking theoretical guarantees of optimality under constrained step budgets.

    In this work, we propose a principled framework for searching the \textbf{O}ptimal \textbf{S}tepsize in diffusion \textbf{S}ampling (\textbf{OSS}), compatible with arbitrary direction strategies. Specifically, we regard the stepsize searching problem as a knowledge distillation problem, where a "student" sampling process with limited steps aims to approximate the results of a "teacher" sampling process with abundant steps. Crucially, we identify that this distillation objective exhibits recursive substructure—the optimal step schedule for $N$ steps inherently contain optimal sub-schedules for fewer steps. Exploiting this insight, we develop a dynamic programming algorithm that systematically derives theoretically optimal step sequences, ensuring maximal proximity to the teacher model's output under any pre-defined student sampling steps. It guarantees that the distilled schedule achieves the closest approximation to the teacher's continuous trajectory under computational constraints.

    Importantly, our experimental analysis demonstrates the universal applicability and robustness of the proposed stepsize distillation framework. It exhibits consistent performance across various model architectures (e.g., U-Net~\cite{edm} vs. Transformer-based~\cite{dit}), ODE solver orders (1st to 3th-order~\cite{dpm++}), noise schedules (linear~\cite{ddim}/Flow Matching~\cite{rf, lipman2022flow}/EDM~\cite{edm}), and diverse denoising direction strategies~\cite{unipc, dpm, dpm++}. Furthermore, when applying the optimal stepsize to text2image and text2video diffusion pipelines, our method achieves \textbf{10$\times$} speedups while preserving \textbf{99.4\%} of the teacher model's performance metrics on GenEval~\cite{ghosh2023geneval} benchmark. Such universal acceleration capability suggests broad practical potential for deploying latency-efficient diffusion models without sacrificing output quality.

    Above all, our key contributions are as follows:

    1. Theoretically Optimal Stepsize Distillation: We formulate stepsize optimization as a dynamic programming problem, proving that our solution can achieve global error minimization.

    2. Architecture-Agnostic Robustness: We demonstrate that the distilled schedules generalize across datasets, noise schedules, and ODE solvers, outperforming heuristic baselines in diverse settings.

    3. Efficient Adaptation: Our method enables lightweight schedule calibration across tasks, achieving 5-10$\times$ speedup with negligible performance degradation.

    \section{Related Work} 

    \subsection{Diffusion sampling direction strategy}

    Recent advances focused on correcting high-order direction for diffusion sampling acceleration. The earliest work~\cite{ddim} adopted first-order Euler for discretization diffusion ODEs.  EDM~\cite{edm} pioneers adopting Heun's\cite{heun} method for second-order gradient estimation between steps, reducing the Euler approximation error. PNDM~\cite{liu2022pseudo} figured out the way of solving the denoising differential equation on manifold. AMED~\cite{zhou2024fast} utilized an additional network to predict the middle point of each step in the second order Heun solver. DPM-Solver~\cite{dpm} leveraged the semi-linear structure of diffusion ODEs, decomposing backward processes into linear/nonlinear components through Taylor expansion. Building on this, DPM-Solver++~\cite{dpm++} introduced a multistep gradient reuse and shifts schedule, further saving the computational efforts. UniPC~\cite{unipc} generalized these via a predictor-corrector framework using Lagrange polynomial interpolation, enabling arbitrary-order accuracy with minimal computational overhead. These innovations collectively established theoretical foundations for balancing discretization precision and computational overhead in diffusion sampling.

    \subsection{Diffusion sampling stepsize strategy}
    Traditional diffusion models typically employ predefined fixed-step strategies. For instance, DiT~\cite{dit} and SDXL~\cite{sdxl} adopt uniform timestep allocations, while EDM~\cite{edm} and CM~\cite{cm} rely on manually crafted schedules. These strategies often yield suboptimal results in few-step sampling scenarios, prompting recent studies to explore stepsize optimization.

    Existing methods fall into two categories: training-free optimization and learning-based strategies. The former approaches, including DM~\cite{dm}, which minimizes inference error bounds via constrained trust-region methods, have a poor generalization ability since it shares the stepsize strategy across different diffusion networks and different solvers. Similarly, AYS~\cite{ays} optimizes stepsize through KL divergence upper bounds (KLUB) but remains confined to SDE frameworks. \cite{jolicoeur2021gotta} gets a better stepsize according to the similarity with the results of the high-order solver, which mixes the direction and stepsize of denoising. \cite{watson2021learning} selects the sampling steps based on dynamic programming and minimizes the sum of ELBO but incurs significant performance degradation. GITS~\cite{gits} also employs dynamic programming to optimize stepsize but exhibits deviations from minimizing global error due to fixed cost matrices. The latter category, exemplified by LD3~\cite{ld3}, leverages a monotonic network for stepsize prediction, which is supervised by global errors. However, it requires additional high computational training. GGDM~\cite{watson2021learning2} adds trainable parameters to the schedule and achieves high-quality samples through well-designed training. We summarize some recent methods in Table \ref{table:related-work} from different perspectives.

\begin{table}[t]
    \setlength{\tabcolsep}{1pt}
    \vspace{-0.1cm}
    \centering
    \begin{tabular}{cccccc} 
    \toprule
    Criterion                                            & \textcolor[rgb]{0.2,0.2,0.2}{AYS~\cite{ays}}                           & \textcolor[rgb]{0.2,0.2,0.2}{GITS~\cite{gits}}                          & \textcolor[rgb]{0.2,0.2,0.2}{DM~\cite{dm}}                           & \textcolor[rgb]{0.2,0.2,0.2}{LD3~\cite{ld3}}                           & \textcolor[rgb]{0.2,0.2,0.2}{OSS}                           \\ 
    \hline
    \textcolor[rgb]{0.2,0.2,0.2}{Training-free} & \textcolor[rgb]{0.2,0.2,0.2}{\checkmark} & \textcolor[rgb]{0.2,0.2,0.2}{\checkmark} & \textcolor[rgb]{0.2,0.2,0.2}{\checkmark} & \textcolor[rgb]{0.2,0.2,0.2}{\ding{53}}   & \textcolor[rgb]{0.2,0.2,0.2}{\checkmark}  \\
    \textcolor[rgb]{0.2,0.2,0.2}{Stability}              & \textcolor[rgb]{0.2,0.2,0.2}{\ding{53}}   & \textcolor[rgb]{0.2,0.2,0.2}{\checkmark} & \textcolor[rgb]{0.2,0.2,0.2}{\checkmark} & \textcolor[rgb]{0.2,0.2,0.2}{\checkmark} & \textcolor[rgb]{0.2,0.2,0.2}{\checkmark}  \\
    \textcolor[rgb]{0.2,0.2,0.2}{Solver-aware}             & \textcolor[rgb]{0.2,0.2,0.2}{\checkmark} & \textcolor[rgb]{0.2,0.2,0.2}{\ding{53}}   & \textcolor[rgb]{0.2,0.2,0.2}{\ding{53}}   & \textcolor[rgb]{0.2,0.2,0.2}{\checkmark} & \textcolor[rgb]{0.2,0.2,0.2}{\checkmark}  \\
    \textcolor[rgb]{0.2,0.2,0.2}{Global error}  & \textcolor[rgb]{0.2,0.2,0.2}{\ding{53}}   & \textcolor[rgb]{0.2,0.2,0.2}{\ding{53}}   & \textcolor[rgb]{0.2,0.2,0.2}{\ding{53}}   & \textcolor[rgb]{0.2,0.2,0.2}{\checkmark}   & \textcolor[rgb]{0.2,0.2,0.2}{\checkmark}  \\
    \bottomrule
    
    \end{tabular}
    \vspace{-0.1cm}
\caption{Related work comparison. Training-free denotes whether eliminating backpropagation during optimization. Stability denotes whether requires specialized stabilization techniques. Solver-aware means the algorithm integrates solver dynamics and network states. Global error means optimizing the final calibration error. }
    \label{table:related-work}
    \end{table}

    \section{Method}
    \subsection{Diffusion sampling preliminaries}

    The continuous diffusion model progressively adds Gaussian noise to a data point $\bm{x_0}$ in the forward process:
    \begin{equation}
        \label{eq:forward}
        q(\bm{x_t}|\bm{x_0}) = \mathcal N(\bm{x_t} |\alpha_t \bm{x_0}, \sigma_t^2 \bm{I})
    \end{equation}
    where $\lambda_t = \frac{\alpha_t^2}{\sigma_t^2}$ denotes the Signal-to-Noise-ratio (SNR). The denoising process generates images from either Stochastic Differential Equation (SDE) solvers~\cite{ho2020denoising, nichol2021improved} or Ordinary Differential Equation(ODE) solvers~\cite{rf,ddim}.
    The key difference between them is whether there is randomness in the generation process. 
    Since ODE eliminates randomness, it often achieves significantly better results with fewer sampling steps.
    Therefore, this paper mainly focuses on the ODE solvers that follows:
    
    \begin{equation}
        \label{eq:ode}d\bm{x_t}=[f_t\bm{x_t}-\frac{1}{2}g^{2}_{t}\nabla_{\bm{x}}logp_{t}(\bm{x_t})]dt
    \end{equation}
    where $f_t=\frac{dlog\alpha_t}{dt}$ and $g_t^2=\frac{d\sigma_t^2}{dt} - 2\frac{dlog\alpha_t}{dt}\sigma_t^2$. The generation process essentially involves solving the ODE and integrating the following expression from timestep $T$ to $0$:
    \begin{equation}
        \label{eq:integral}
        \bm{x_{0}}= \bm{x_{T}}+ \int_{T}^{0} f_t\bm{x_t}-\frac{1}{2}g^{2}_{t}\nabla_{\bm{x}}logp_{t}(\bm{x_t}) dt
    \end{equation}

    In practical applications, we usually discretize it into a finite number of $N$ steps to speed up the inference process:

    \begin{equation}
        \label{eq:discretize}
        \bm{x_{0}}= \bm{x_{T}}+ \sum_{i=N}^{1}\bm{v_{\theta,i}} (t_{i-1} - t_{i})
    \end{equation}
    where $\bm{x_T} \sim N(0,1)$ is the initial noise, $\bm{\theta}$ denotes the parameter of denoising model, $\bm{v_{\theta,i}}$ denotes the optimization direction on step  $i$. For simplicity, we use $\bm{v_{\theta}}$ to represent the direction strategy. $t_{i-1} - t_{i}$ denotes the stepsize on each optimization step. Many previous works have explored the selection of optimization direction $\bm{v_{i}}$. For example, in Heun~\cite{heun,edm}, $\bm{v_i} = \frac{1}{2}(\bm{v}_{t_i} + \bm{v}_{t_{i+1}})$, 
    in DPM-Solver~\cite{dpm}, $\bm{v_i} = \bm{v}_{t_\lambda(\frac{\lambda_{t_{i}} + \lambda_{t_{i+1}}}{2})}$, $t_\lambda$ is the inverse function mapping from $\lambda$ to $t$.

    However, few works have discussed the selection of stepsize $t_i$. 
    Most of the previous works can be divided into two categories. The first class selects a uniform $t_i$:
    \begin{equation}
        t_i = \frac{i}{N} \quad where\ i = 0 \ldots N
    \end{equation}
    such practices include LDM~\cite{ldm}, DiT~\cite{dit}, FLUX~\cite{flux}, SDXL~\cite{sdxl}, etc. The second class includes EDM~\cite{edm}, CM~\cite{cm}, which adopts a handcraft stepsize schedule:

    \begin{equation}
        \label{eq:edm}t_{i}= (t_{1}^{1/\rho}+ \frac{i-1}{N-1}(t_{N}^{1/\rho}- t_{1}
        ^{1/\rho}))^{\rho}
    \end{equation}
    However, these ad-hoc schedules always lead to a sub-optimal result. In this paper, we emphasize that the choice of stepsize is particularly crucial, especially in scenarios with limited steps. Besides, the stepsize operates orthogonally to the directional strategy.

    \subsection{Definition of optimal denoising stepsize}

    Although discretizing the ODE trajectory can significantly increase the speed, it will deviate from the original ODE and cause the results to deteriorate to a certain extent. Therefore, the stepsize discretization aims to generate results that are as consistent as possible with the non-discretization. This is similar to traditional distillation, so we formulate our task as stepsize distillation with the difference being that no parameter training is required:

    Suppose that we have a teacher inference schedule, including both the direction schedule ($\bm{v_{\theta}}$ in Equation \ref{eq:discretize}) and the stepsize schedule ($\{t^N\} = \{t_{N},t_{N-1},...,t_{1},t_{0}\}$), where the number of steps $N$ is large enough to approximate infinite steps sampling results. We abbreviate the result generated from the initial point $\bm{x_T}$ through the stepsize strategy $\{t^N\}$ and direction strategy $\bm{v_\theta}$ according to Equation \ref{eq:discretize} as $\mathcal{F}(\bm{x_T},\bm{v_\theta},\{t^N\})$.
    For one step denoising of state $\bm{x_i}$($t_j < t_i$) starts from $t_i$ and ends at $t_j$, the results can be format as follows: 
    \begin{equation}
        \label{eq:onestep}
        \mathcal{F}(\bm{x_i}, \bm{v_\theta}, \{t_i,t_j\}) = \bm{x_i} + \bm{v_\theta} (t_j - t_i)
    \end{equation}

     We aim to distill a student stepsize schedule,
    ($\{s^M\} = \{s_{M},s_{M-1},...,s_{1},s_{0}\}$), where $\{s^M\}$ is a subset of $\{t^N\}$ and $M$ is smaller than $N$. 
    For a fixed initial noise $\bm{x_T}$, the optimal student stepsize schedule $\{s^M\}^o$ aims to minimize the global discretization error with the teacher generation result:

    \begin{equation}
        \{s^M\}^o = \mathop{\arg\min}_{\{s^M\} \subset \{t^N\}}\norm{ \mathcal{F}(\bm{x_T},\bm{v_\theta},\{t^N\}) - \mathcal{F}(\bm{x_T},\bm{v_\theta},\{s^M\}) }^{2}_2
    \end{equation}

    \subsection{Dynamic programming based calibration}

    \subsubsection{Recursive subtasks}
    Our goal is to use $M$ student steps to approximate $N$ teacher steps sampling results as closely as possible. We separate it into subtasks: \emph{using $i$ student steps to approximate $j$ teacher steps sampling results}.

    Therefore, we make such notations: the sampling trajectory of the teacher model is $\{\bm{X^t}\} = \{\bm{x}[t_N],\bm{x}[t_{N-1}],...,\bm{x}[t_1],\bm{x}[t_0]\}$ based on the stepsize schedule $\{t^N\}$, where $\bm{x}[t_j]=\mathcal{F}(\bm{x_T},\bm{v_\theta},\{t_N,...,t_j\})$ denotes the sampling result at timestep $t_j$. For simplicity, we abbreviate $\bm{x}[t_j]$ as $\bm{x}[j]$. For the student model, we denote the optimal denoised result using $i$ step to approximate $\bm{x}[j]$ as $\bm{z}[i][j]$. Moreover, We additionally record that $\bm{z}[i][j]$ comes from timestep $r[j]$($r[j] > j$), which means:
    \begin{equation} \label{eq:dp1}
        r[j] = \mathop{\arg\min}_{j+1,...,N} \norm{\bm{x}[j] - \mathcal{F}(\bm{z}[i-1][r[j]], \bm{v_\theta}, \{r[j],j\}) }^{2}_2
    \end{equation}

    We can find that these subtasks demonstrate a Markovian recursive property contingent upon the number of student steps utilized: 
    The student model's $i$-th step approximation for the teacher's $j$-th step is determined by its prior result, where the approximation at timestep $r[j]$ utilized $i-1$ denosing steps. It can be formally expressed as:

    \begin{equation} \label{eq:dp2}
        \bm{z}[i][j] = \mathcal{F}(\bm{z}[i-1][r[j]], \bm{v_\theta}, \{r[j],j\})
    \end{equation}

    By combining Equation ~\ref{eq:dp2} and Equation ~\ref{eq:onestep}, the recursion relation can be written as:

    \begin{equation} \label{eq:dp3}
        \bm{z}[i][j] =  \bm{z}[i-1][r[j]] + \bm{v_{\theta}}(j-r[j])
    \end{equation}

\begin{figure}
        \centering
        \includegraphics[width=0.5\textwidth]{
            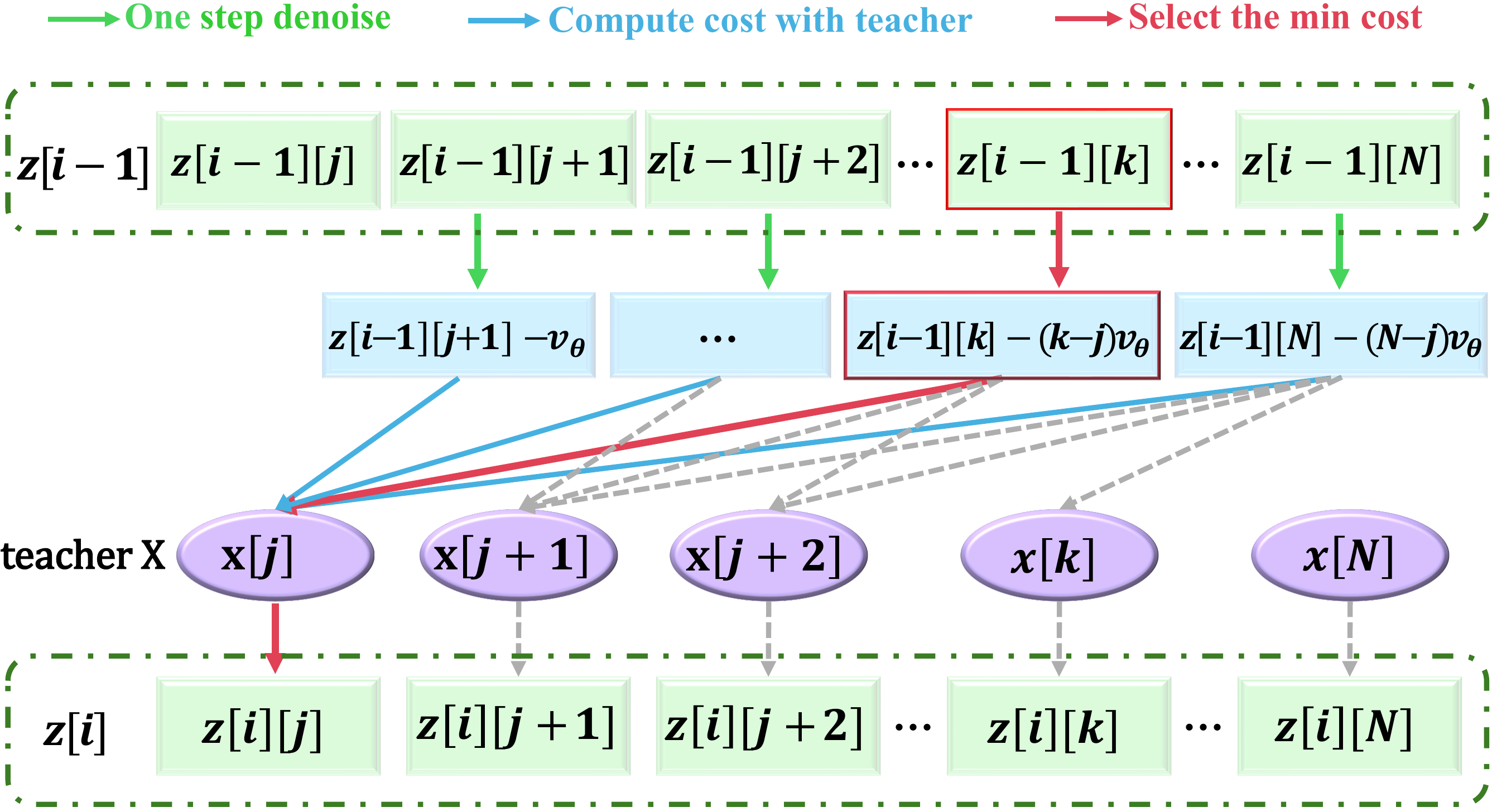
        }
        \vspace{-0.5cm}
        \caption{Subtask illustration of the recursive subtasks. The optimal results at timestep $j$ using $i$ step denosing ($z[i][j]$) derives from the $i-1$ step optimal denosing results($z[i-1]$).}
        \label{fig:method}
    \end{figure}

    The illustration of solving the recursive subtasks is shown in Figure \ref{fig:method}. 

    \begin{algorithm}[h]
        \caption{Optimal Stepsize Distillation.}
        \label{alg:algorithm1}
        \begin{algorithmic}
            [1]
            {\Statex $\bm{D_{\theta}}:$ Denoising model which outputs the velocity.
            \Statex $\bm{x}[N]$, $\bm{z}$ samples from $ \mathcal{N}(0,1)$} \\

            \textcolor{gray}{\# get teacher trajectory.}
            \For{$\bm{i}$ = $N$ to 1}

            \State{$\bm{v}$ = $\bm{D_{\theta}}(\bm{x}[i], i$)} 
            \State{$\bm{x}[i-1]$ = $\bm{x}[i]$ - $\bm{v}$}
            \EndFor \\
            
            \textcolor{gray}{\# use the student i steps result to approach the teacher's j steps result. }
            \For{$\bm{i}$ = 1 to $M$}  
            \State{$\bm{v}$ = $\bm{D_{\theta}}(\bm{z}[i-1,:],range(N,1))$}
            \State{$\bm{z_{tmp}}$ = ones($N+ 1, N$)$\times$Inf} 
            \For{$\bm{j}$ in $N$ to 1}
                \State{ $\bm{z_{tmp}}[j,:]$ = $\bm{z}[i-1,j] - \bm{v}[j]$$\times range(0,j-1)$}
            \EndFor 
            \For{$\bm{j}$ in $N- 1$ to 0}
            \State{$\bm{cost}$ = $\bm{MSE}(\bm{z_{tmp}}[:,j], \bm{x}[j])$}
            \State{$\bm{r}[i,j]$ = $argmin(\bm{cost})$} 
            \State{$\bm{z}[i,j]$ = $\bm{z_{tmp}}[\bm{r}[i,j],j]$}
            \EndFor 
            \EndFor \\
            \textcolor{gray}{\# retrieve the optimal trajectory. }
            \State{$\bm{R}$ = [0]}
            \For{$i$ in $M$ to 1} 
            \State{$j$ = $\bm{R}[-1]$}

            \State{$\bm{R}$.append$(\bm{r}[i][j])$} 
            \EndFor 
            \State {\bfseries return} $\bm{R}[1:]$
        \end{algorithmic}
    \end{algorithm}

    \subsubsection{Dynamic programming}
    According to the recursion formula, we use dynamic programming to solve the stepsize selection problem for the student model. This process can be regarded as searching $\bm{z}[i][j]$ and recording each $r[j]$ as $i$ increases. Then we can obtain the final result of using $M$ student steps approximating $N$ teacher steps $\bm{z}[M][N]$. Finally, we backtrack the timestep $r[j]$ as the optimal trajectory. In this way, we distilled the optimal stepsize of the student model from the teacher steps. The pseudo code is provided in \Cref{alg:algorithm1}.

    \subsubsection{Theoretically optimal stepsize}
    To rigorously establish the theoretical optimality of the dynamic programming algorithm in attaining the optimal stepsize schedule, we demonstrate that this subtask decomposition simultaneously satisfies both \emph{ovealapping subproblems} and \emph{optimal substructure}. 

    {\bf Overlapping Subproblems}. This can be easily found since each denoised step is only dependent on the previous one step result, which is independent of the denoised history before. Take the one order solver as an example. Suppose we want to denoise from timestep $s$ to timestep $t$, we have:
    \begin{equation}
        \bm{x_t} = \bm{x_s} + \bm{v_{\theta}}(t-s)    
    \end{equation}
    One special case is the multistep high order solvers such as DPM-Solver~\cite{dpm++} and UniPC~\cite{unipc}. 
    Compared with one order solver, the only difference is that the denoising process needs to rely on multi-step results. In this case, the overlapping subproblem changes into denoising from multiple states with different noise intensities. We leave the details about searching with high order solvers in the Appendix.

    {\bf Optimal Substructure}. 
    The optimal substructure enables pruning suboptimal denoising trajectories in advance by dynamic programming. Here, we illustrate this substructure in \Cref{lemma} and leave the proof in the Appendix.

    \begin{lemma} \label{lemma}
        The optimal m step denosing results of student solver $z[m]$ always derives from the optimal m-1 step results $z[m-1]$ with additional one step denosing.
    \end{lemma}

    \begin{figure}[H]
        \centering
        \vspace{-0.6cm}
        \includegraphics[width=0.5\textwidth]{
            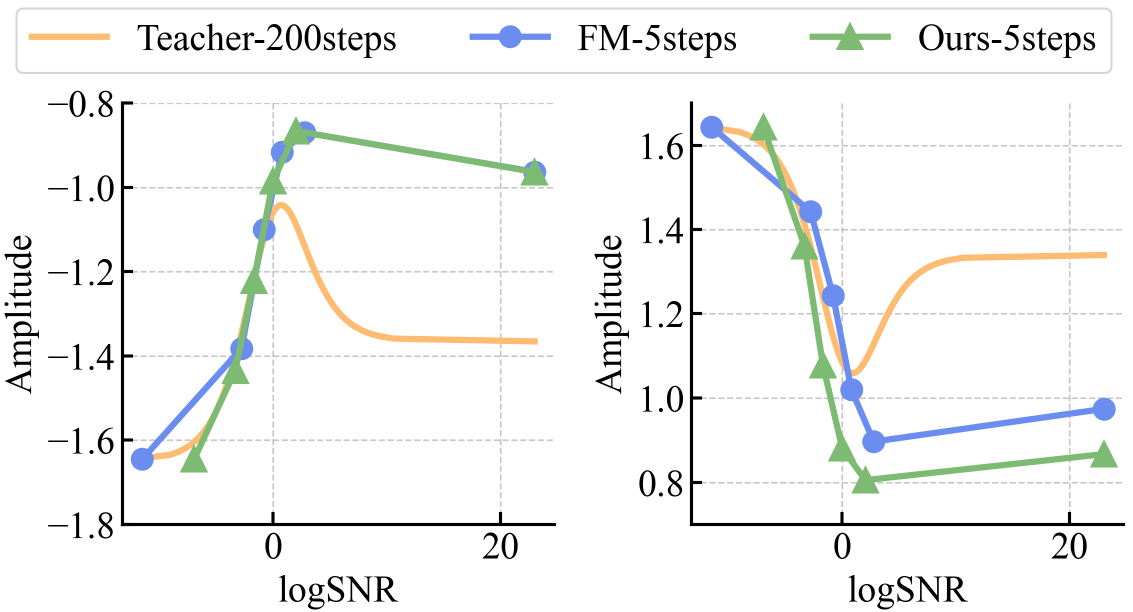
        }
        \vspace{-0.6cm}
        \caption{Amplitude of input tensor throughout denoising steps. The left and right plot the quantile of 5\%, and 95\% respectively.}
        \label{fig:maxmin}
    \end{figure}

    \subsection{Amplitude calibration} \label{sec:renorm}

    While our step optimization enables efficient trajectory approximation, we identify a critical challenge when applying few-step inference: systematic amplitude deviation becomes pronounced in low-step regimes. 
    In Figure \ref{fig:maxmin}, we plot the the quantile of 5\% and 95\%  percentiles of the input tensor to denoising models across varying noise intensities.
    We find that the distribution changes significantly between the few-step (5) and multiple-step (200) trajectories, especially in the later stages of the denoising process.

    To address this, we propose a per-step affine transformation that aligns student outputs with teacher amplitude characteristics. For each noise level $\lambda_t$, we calibrate the transformation using a linear function:
    
    \begin{equation} 
    \bm{\hat{x}}_t = \frac{S_{95}-S_5}{x_{t,95}-x_{t,5}}  \bm{x}_t
    \end{equation}
    
    where $S_{95}$ and $S_5$ denote the quantile ranges of 95\%-5\% of the teacher on 512 samples, which are calculated in advance. $x_{t,95}$ and $x_{t,5}$ are the quantile ranges of 95\% - 5\% of the student at step $t$ during denosing. In \Cref{sec:amplitude}, we show that applying this calibration at the test time can significantly improve details such as color and texture.

    \section{Experiments}
    
        \textbf{Implementation details.} 
    Our algorithm features plug-and-play compatibility, allowing for seamless integration into  various diffusion frameworks with minimal adaptation effort. To ensure consistent parameterization across different implementations, we adopt the velocity prediction paradigm from Flow Matching(FM)\cite{rf, lipman2022flow}  as our unified inference target. This requires wrapping the existing models with two standardization operations: (1) reparameterize the prediction target to velocity space, and (2) align the sampling trajectory based on the Signal-to-Noise ratio (SNR) matching. As theoretically guaranteed by~\cite{improverectifiedflow}, this standardization does not affect the sampling result while enabling unified stepsize optimization across architectures. In the Appendix, we will explain this unification in detail.
    
    \textbf{Evaluation Protocol}
    We prioritize minimizing the divergence of the trajectory from the teacher models as our main optimization objective. For quantitative evaluation, we employ PSNR to measure pixel-level fidelity between student and teacher output. Unless otherwise specified, all teacher models utilize 200 sampling steps to empirically balance computational efficiency with output quality across most test domains.

    \subsection{Robustness analysis}
    We conduct comprehensive ablation studies to validate our method's robustness across four critical dimensions: noise schedules (\Cref{sec:teacher_schedule}), number of teacher steps (\Cref{sec:teacher_steps}), ODE solver orders (\Cref{sec:teacher_order}), and diverse modeling frameworks (\Cref{sec:different_method}).

    \subsubsection{Robustness to noise schedules} \label{sec:teacher_schedule}
    To evaluate schedule invariance, we performed class-conditional image generation experiments on ImageNet-64\cite{russakovsky2015imagenet} using an EDM-based~\cite{edm} diffusion model. We test three established noise schedules as teacher configurations: 
    \begin{itemize} 
    \item EDM's exponential noise schedule~\cite{edm}. 
    \item DDIM's linear noise schedule~\cite{ddim}. 
    \item Flow Matching's schedule~\cite{lipman2022flow}  
    \end{itemize}
    Each teacher generates reference samples using 200 steps. We compare our optimized steps against naive uniform step reduction, measuring similarity through PSNR between student and teacher outputs. As shown in \Cref{table:pixel_teacherschedule}, our method achieves 2.65 PSNR improvements on average over baseline step reduction across all schedules, demonstrating consistent performance regardless of the teacher's schedule.

    \begin{table}[h]
    \setlength{\tabcolsep}{2pt}

\centering
\begin{tabular}{cc|ccc} 
\toprule
                                                        & \textcolor[rgb]{0.2,0.2,0.2}{steps} & \textcolor[rgb]{0.2,0.2,0.2}{EDM}                                                          & \textcolor[rgb]{0.2,0.2,0.2}{DDIM}                                                         & \textcolor[rgb]{0.2,0.2,0.2}{FM}                                                            \\ 
\hline
\multirow{4}{*}{\begin{tabular}[c]{@{}c@{}}\textcolor[rgb]{0.2,0.2,0.2}{Naive step}~\\\textcolor[rgb]{0.2,0.2,0.2}{reduction}\end{tabular}}

& \textcolor[rgb]{0.2,0.2,0.2}{50}  & \textcolor[rgb]{0.2,0.2,0.2}{33.71}                                                        & \textcolor[rgb]{0.2,0.2,0.2}{34.31}                                                        & \textcolor[rgb]{0.2,0.2,0.2}{31.78}                                                         \\
                                                        & \textcolor[rgb]{0.2,0.2,0.2}{20}  & \textcolor[rgb]{0.2,0.2,0.2}{25.76}                                                        & \textcolor[rgb]{0.2,0.2,0.2}{26.12}                                                        & \textcolor[rgb]{0.2,0.2,0.2}{23.97}                                                         \\
                                                        & \textcolor[rgb]{0.2,0.2,0.2}{10}  & \textcolor[rgb]{0.2,0.2,0.2}{20.83}                                                        & \textcolor[rgb]{0.2,0.2,0.2}{20.89}                                                        & \textcolor[rgb]{0.2,0.2,0.2}{19.00}                                                         \\
                                                        & \textcolor[rgb]{0.2,0.2,0.2}{5}   & \textcolor[rgb]{0.2,0.2,0.2}{16.38}                                                        & \textcolor[rgb]{0.2,0.2,0.2}{15.96}                                                        & \textcolor[rgb]{0.2,0.2,0.2}{15.69}                                                         \\ 
\hline
\multirow{3}{*}{\textcolor[rgb]{0.2,0.2,0.2}{OSS}}     & \textcolor[rgb]{0.2,0.2,0.2}{20}  & \textcolor[rgb]{0.2,0.2,0.2}{28.21(\textbf{+}}\textcolor[rgb]{0.2,0.2,0.2}{\textbf{2.45})} & \textcolor[rgb]{0.2,0.2,0.2}{28.07(\textbf{+}}\textcolor[rgb]{0.2,0.2,0.2}{\textbf{1.95})} & \textcolor[rgb]{0.2,0.2,0.2}{28.51(\textbf{+}}\textcolor[rgb]{0.2,0.2,0.2}{\textbf{4.54})}  \\
                                                        & \textcolor[rgb]{0.2,0.2,0.2}{10}  & \textcolor[rgb]{0.2,0.2,0.2}{22.88(\textbf{+2}}\textcolor[rgb]{0.2,0.2,0.2}{\textbf{.05})} & \textcolor[rgb]{0.2,0.2,0.2}{22.79(\textbf{+}}\textcolor[rgb]{0.2,0.2,0.2}{\textbf{1.9})}  & \textcolor[rgb]{0.2,0.2,0.2}{23.02(\textbf{+}}\textcolor[rgb]{0.2,0.2,0.2}{\textbf{4.02})}  \\
                                                        & \textcolor[rgb]{0.2,0.2,0.2}{5}   & \textcolor[rgb]{0.2,0.2,0.2}{18.41(\textbf{+}}\textcolor[rgb]{0.2,0.2,0.2}{\textbf{2.03})} & \textcolor[rgb]{0.2,0.2,0.2}{18.05(\textbf{+}}\textcolor[rgb]{0.2,0.2,0.2}{\textbf{2.09})} & \textcolor[rgb]{0.2,0.2,0.2}{18.53(\textbf{+}}\textcolor[rgb]{0.2,0.2,0.2}{\textbf{2.84})}  \\
\bottomrule
\end{tabular}
\vspace{-0.3cm}
        \caption{ImageNet-64 pixel space conditional image generation with different teacher schedules. All the student solvers are learned from the corresponding 200-step teacher schedule. Results are shown in PSNR calculated with 200-step corresponding teacher results.}
        \label{table:pixel_teacherschedule}
        \vspace{-0.1cm}
\end{table}

    \subsubsection{Robostness to number of steps} \label{sec:teacher_steps}
    We analyze the impact of the number of teacher steps based on ImageNet experiments. The teacher solver adopts the DDIM schedule and generates images using various numbers of steps from 100 to 1000 (student fixed at 20 steps). We measure the PSNR against 1000-step teacher results. \Cref{table:diff_tea} reveals that the performance basically converged in 200 steps. This indicates that the 200-step teacher model already provides an adequate search space for the dynamic programming algorithm to identify the optimal 20-step policy. Moreover, our method also maintains fidelity well, even when the teacher model has only 100 steps.

\begin{figure}[t]
        \centering
        \includegraphics[width=0.45\textwidth]{
            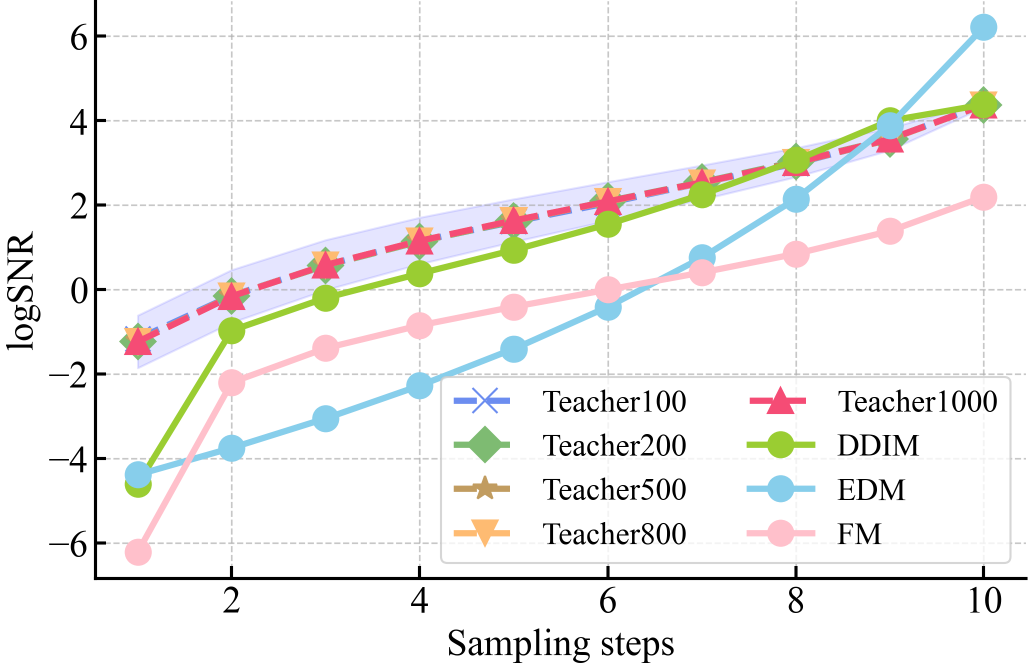
        }
        \caption{Step schedule for different solvers. Our method achieves nearly identical results from different teacher model steps.}
        \label{fig:step_variance}
    \end{figure}

    We further analyze the variance of step distribution across samples in \Cref{fig:step_variance}. For teacher solvers with different steps, the optimal step sequences almost overlap. Besides, different samples demonstrate minor differences (light purple area in the figure). This allows us to generalize to unsearched samples by averaging over a selected subset of instances, thereby reducing the inference overhead in a zero-shot manner. We named this setting as \emph{OSS-ave}. Specifically, by employing 512 samples to compute the mean of the optimal stepsize schedule, we achieve results that closely approximate those obtained through instance-specific optimization, as shown in \Cref{table:diff_tea}.

\begin{table}[htbp]

\centering
\begin{tabular}{cccccc} 
\toprule
Teacher steps                                & \textcolor[rgb]{0.2,0.2,0.2}{100}   & \textcolor[rgb]{0.2,0.2,0.2}{200}   & \textcolor[rgb]{0.2,0.2,0.2}{500}   & \textcolor[rgb]{0.2,0.2,0.2}{800}   & \textcolor[rgb]{0.2,0.2,0.2}{1000}   \\ 
\hline
\textcolor[rgb]{0.2,0.2,0.2}{DDIM teacher} & \textcolor{gray}{37.97} & \textcolor{gray}{44.24} & \textcolor{gray}{52.93} & \textcolor{gray}{57.30} & \textcolor{gray}{-}      \\
\hline
\textcolor[rgb]{0.2,0.2,0.2}{OSS} & \textcolor[rgb]{0.2,0.2,0.2}{27.04} & \textcolor[rgb]{0.2,0.2,0.2}{28.07} & \textcolor[rgb]{0.2,0.2,0.2}{27.11} & \textcolor[rgb]{0.2,0.2,0.2}{27.12} & \textcolor[rgb]{0.2,0.2,0.2}{27.13}  \\
\textcolor[rgb]{0.2,0.2,0.2}{OSS-ave}  & \textcolor[rgb]{0.2,0.2,0.2}{26.41} & \textcolor[rgb]{0.2,0.2,0.2}{26.47} & \textcolor[rgb]{0.2,0.2,0.2}{26.47} & \textcolor[rgb]{0.2,0.2,0.2}{26.48} & \textcolor[rgb]{0.2,0.2,0.2}{26.48}  \\
\bottomrule
\end{tabular}

\caption{ImageNet-64 pixel space conditional image generation results with different teacher steps. All the teacher solver leverages DDIM schedules. Students adopt 20-steps for inference. Results are shown in PSNR calculated with the 1000-step teacher.}
\label{table:diff_tea}
\end{table}

    \subsubsection{Robustness to higher order solver} \label{sec:teacher_order}

\begin{figure}[H]
        \centering

        \includegraphics[width=0.5\textwidth]{
            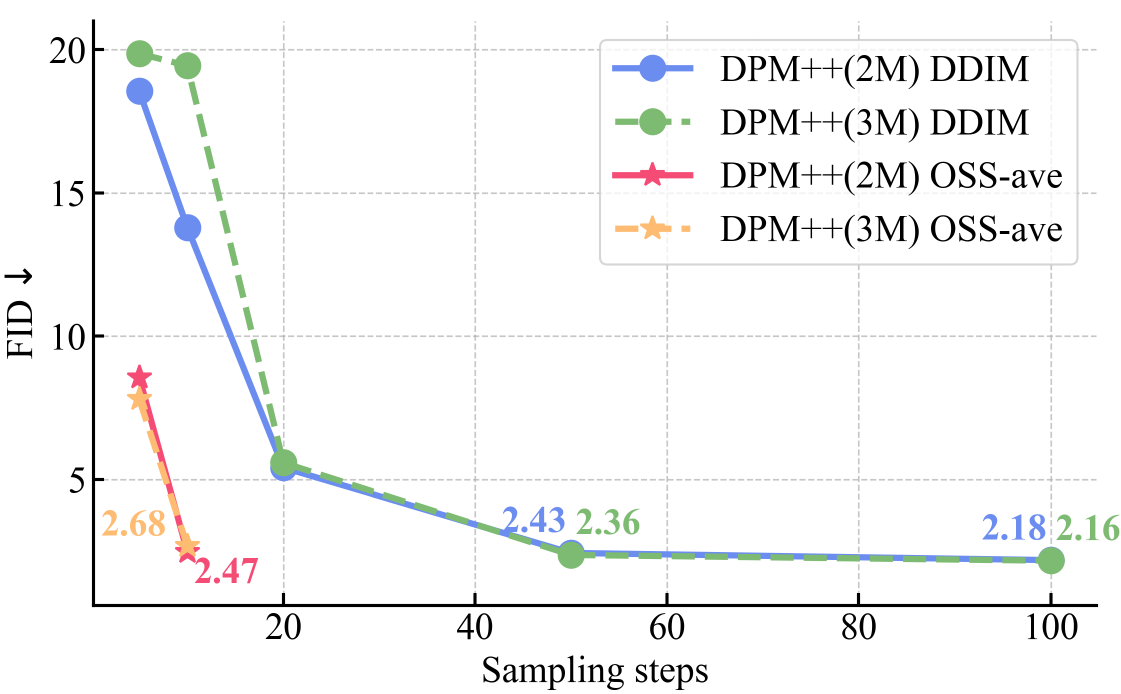
        }
        \caption{ImageNet-256 latent space conditional image generation results using different order solvers.}
        \label{fig:order}
    \end{figure}

    Our optimal stepsize searching algorithm can be applied to higher-order solvers. We validate it using DiT-XL/2 model\cite{dit} on ImageNet 256×256 using classifier-free guidance as 1.5. Figure ~\ref{fig:order} demonstrates that our step sequences combined with third-order solvers achieves significant improvements over the baseline DDIM schedule.

    This verifies that optimal step scheduling and the solver order constitute complementary axes for few-step sampling improvement - their synergy enables practical acceleration without quality degradation.

    \subsubsection{Generalization Across frameworks} \label{sec:different_method}
    We validate framework-agnostic applicability through two different extensions:
    \textbf{Masked Autoregressive (MAR) Generation:} By applying the optimal stepsize schedule to the MAR \cite{mar} ImageNet-256 model, we reduce the sampling steps from 500 to 5, achieving 100$\times$ acceleration while maintaining competitive performance (FID=4.67). When decreasing steps to 50, the FID remained nearly unchanged (2.66\textrightarrow2.59), demonstrating robust stability. Note that the model MAR-Base contains 32 tokens, but we only search for the first token and apply the result steps to the other 31 tokens, which further demonstrates the robustness.

    \textbf{Video Diffusion:} In Open-Sora \cite{opensora} frameworks, our optimized schedules enable 10$\times$ speedups while preserving visual fidelity. Visualization results are shown in the Appendix. We find that simply reducing the sampling step to 20 steps completely changes the generated video. However, by adjusting the sampling steps, the generated result is almost the same as the teacher with 10$\times$ speedup.

   \begin{figure}[H]
        \vspace{-0.4cm}
        \centering
        \includegraphics[width=0.5\textwidth]{
            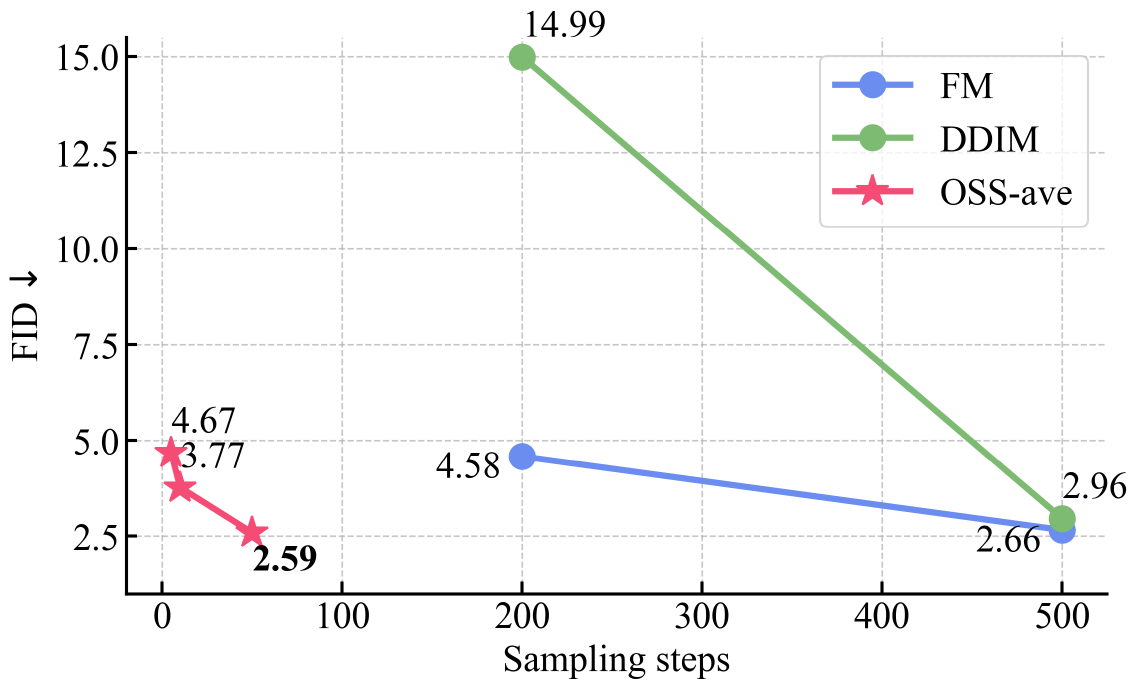
        }
        \vspace{-0.7cm}
        \caption{ImageNet-256 latent space conditional image generation results with MAR-Base.}
 \label{fig:mar}
 
    \end{figure}

    \begin{figure}[t]
        \centering
        \vspace{-0.3cm}
        \includegraphics[width=0.5\textwidth]{
            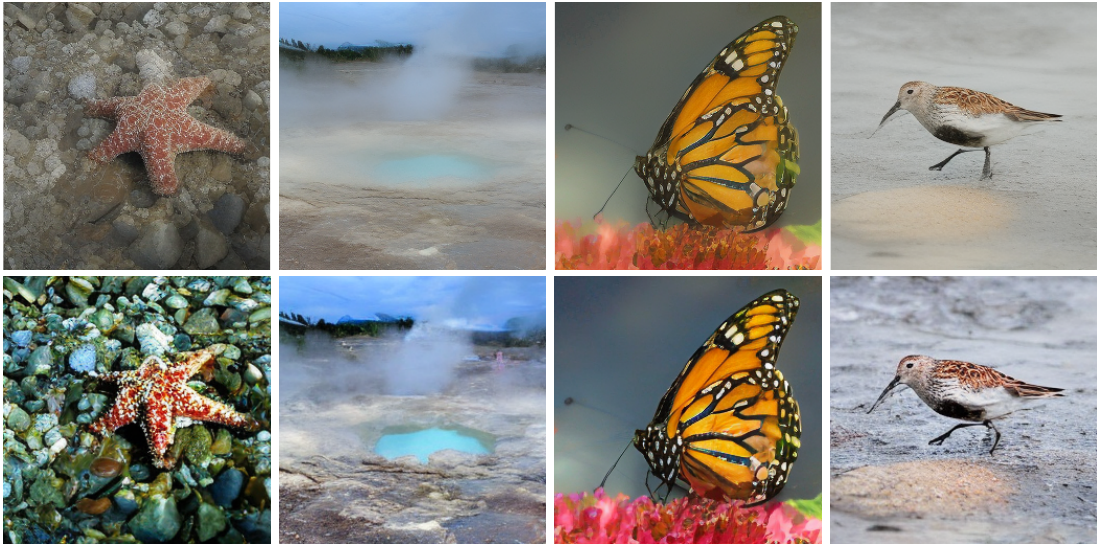
        }
        \vspace{-0.6cm}
        \caption{Through amplitude adjustment, the synthesized outputs (bottom row) exhibit enhanced details.} 
        \label{fig:renorm}
    \end{figure}

    \subsection{Extensions of optimal stepsize}
    \subsubsection{Amplitude calibration} \label{sec:amplitude}
    As demonstrated in \Cref{sec:renorm}, insufficient sampling steps may induce significant amplitude discrepancies between generated intermediate states and the teacher model's trajectories. To investigate it, we conducted quantitative and qualitative analysis on the ImageNet dataset. Although amplitude adjustment may slightly decrease the PSNR compared to the teacher model's outputs ($18.53 \rightarrow 16.92$), the adjusted samples exhibit enhanced structural detail and demonstrate improved realism, as evidenced in ~\Cref{fig:renorm}.

\begin{table}[H]

    \centering
    \begin{tabular}{ccccc} 
    \toprule
                                       & \textcolor[rgb]{0.2,0.2,0.2}{$x_t$-MSE}                                & \textcolor[rgb]{0.2,0.2,0.2}{$x_0$-MSE} & \textcolor[rgb]{0.2,0.2,0.2}{$x_0$-IncepV3} & \textcolor[rgb]{0.2,0.2,0.2}{$x_t$-percep}\textcolor[rgb]{0.2,0.2,0.2}{}  \\ 
    \hline
    \textcolor[rgb]{0.2,0.2,0.2}{PSNR $\uparrow$} & \textcolor[rgb]{0.2,0.2,0.2}{18.5}\textcolor[rgb]{0.2,0.2,0.2}{4} & \textcolor[rgb]{0.2,0.2,0.2}{22.45}                                       & \textcolor[rgb]{0.2,0.2,0.2}{9.91}                                          & \textcolor[rgb]{0.2,0.2,0.2}{18.27}                                           \\
    \bottomrule
    \end{tabular}
    \vspace{-0.3cm}
    \caption{Ablation studies on different functions.}
    \label{table:metrics}
    \end{table}

    \begin{figure*}[t]
        \centering
        \vspace{-0.6cm}
        \includegraphics[width=\textwidth]{
            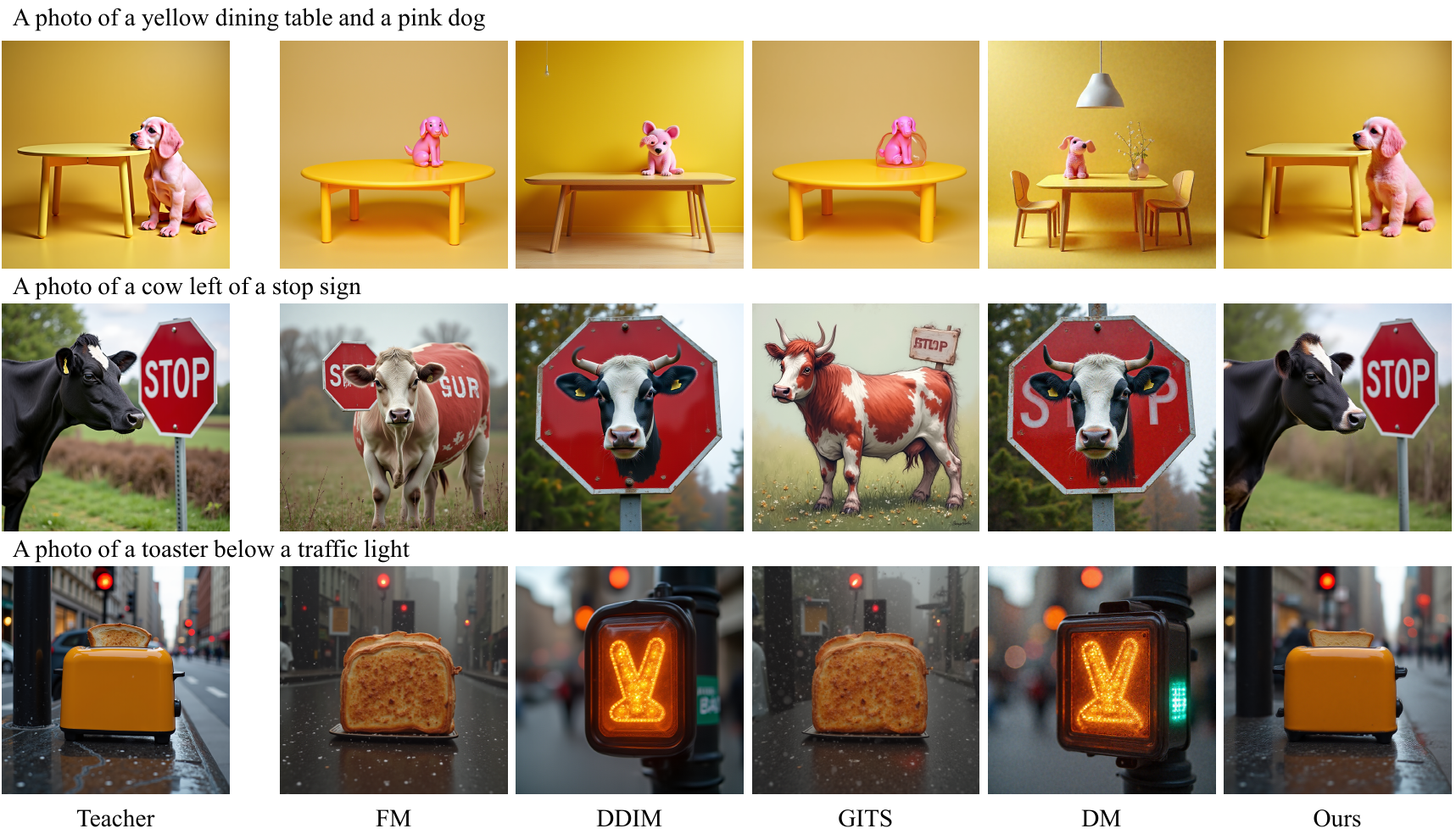
        }
        \vspace{-0.8cm}
        \caption{Visualization results on Geneval benchmark. Our optimal sampling schedule can produce results that are more similar to those of multi-step teachers, inherited strong instruction following ability.} 
        \label{fig:geneval}
    \end{figure*}

    \subsubsection{Metric functions}

    Conventional continuous diffusion alignment relies on MSE between teacher and student intermediate states $x_t$. However, MSE may induce over-smoothing in pixel-level generation tasks. We thus explore three metrics: (1) MSE on reparameterized clean data $x_0$, this can be regarded as SNR-weighted MSE on $x_t$ ~\cite{hang2023efficient}; (2) Inception-V3~\cite{szegedy2016rethinking} feature distance on reparameterized $x_0$; (3) perceptual distance utilizing the denoising network~\cite{dit} on intermediate feature $x_t$.

    We evaluate the three aforementioned metrics on the ImageNet-64 generation with the pretrained EDM model and summarized the results in Table ~\ref{table:metrics}. The $x_0$-MSE setting outperforms the default distance metric calculated on $x_t$. However, the Inception-V3 feature metric performs worse, potentially attributed to lacking generalization ability on noisy data. Meanwhile, using the denoising model to calculate the perceptual distance does not demonstrate significant improvement.

\begin{table}[H]

\vspace{-0.2cm}
\centering

\begin{tabular}{c|cc} 
\toprule
                                                                    & \textcolor[rgb]{0.2,0.2,0.2}{w}\textcolor[rgb]{0.2,0.2,0.2}{/o training} & \textcolor[rgb]{0.2,0.2,0.2}{w}\textcolor[rgb]{0.2,0.2,0.2}{~training}               \\ 
\hline
\textcolor[rgb]{0.2,0.2,0.2}{U}\textcolor[rgb]{0.2,0.2,0.2}{niform} & \textcolor[rgb]{0.2,0.2,0.2}{5}\textcolor[rgb]{0.2,0.2,0.2}{2.73}        & \textcolor[rgb]{0.2,0.2,0.2}{2}\textcolor[rgb]{0.2,0.2,0.2}{4.53}                    \\
\textcolor[rgb]{0.2,0.2,0.2}{O}\textcolor[rgb]{0.2,0.2,0.2}{SS}     & \textcolor[rgb]{0.2,0.2,0.2}{4}\textcolor[rgb]{0.2,0.2,0.2}{1.96}        & \textcolor[rgb]{0.2,0.2,0.2}{\textbf{7.89}}  \\
\bottomrule
\end{tabular}
\vspace{-0.2cm}
\caption{Sampling results on ImageNet-256. Uniform is the schedule adopted in DDIM.}
\label{table:distill}
\end{table}

\begin{table}[H]
\centering
\vspace{-0.2cm}
\centering
\begin{tabular}{c|cc|cc} 
\toprule
\multicolumn{1}{c}{\textcolor[rgb]{0.2,0.2,0.2}{Method}} & \textcolor[rgb]{0.2,0.2,0.2}{5steps}          & \multicolumn{1}{c}{\textcolor[rgb]{0.2,0.2,0.2}{10steps}} & \textcolor[rgb]{0.2,0.2,0.2}{5steps}          & \textcolor[rgb]{0.2,0.2,0.2}{10steps}               \\ 
\hline
                                                         & \multicolumn{2}{c|}{\textcolor[rgb]{0.2,0.2,0.2}{ImageNet-64~\cite{russakovsky2015imagenet}}}                                         & \multicolumn{2}{c}{\textcolor[rgb]{0.2,0.2,0.2}{LSUN-Bed-256~\cite{yu2015lsun}}}  \\ 
\hline
\textcolor[rgb]{0.2,0.2,0.2}{EDM}                        & \textcolor[rgb]{0.2,0.2,0.2}{17.72}          & \textcolor[rgb]{0.2,0.2,0.2}{22.33}                      & \textcolor[rgb]{0.2,0.2,0.2}{15.60}          & \textcolor[rgb]{0.2,0.2,0.2}{19.61}                \\
\textcolor[rgb]{0.2,0.2,0.2}{DDIM}                       & \textcolor[rgb]{0.2,0.2,0.2}{17.69}          & \textcolor[rgb]{0.2,0.2,0.2}{22.22}                      & \textcolor[rgb]{0.2,0.2,0.2}{15.39}          & \textcolor[rgb]{0.2,0.2,0.2}{20.18}                \\
\textcolor[rgb]{0.2,0.2,0.2}{FM}                         & \textcolor[rgb]{0.2,0.2,0.2}{17.80}          & \textcolor[rgb]{0.2,0.2,0.2}{22.47}             & \textcolor[rgb]{0.2,0.2,0.2}{14.43}          & \textcolor[rgb]{0.2,0.2,0.2}{18.23}                \\
\textcolor[rgb]{0.2,0.2,0.2}{DM}                         & \textcolor[rgb]{0.2,0.2,0.2}{17.39}          & \textcolor[rgb]{0.2,0.2,0.2}{20.50}                      & \textcolor[rgb]{0.2,0.2,0.2}{\textbf{18.90}} & \textcolor[rgb]{0.2,0.2,0.2}{20.67}                \\
\textcolor[rgb]{0.2,0.2,0.2}{GITS}                       & \textcolor[rgb]{0.2,0.2,0.2}{16.29}          & \textcolor[rgb]{0.2,0.2,0.2}{19.40}                      & \textcolor[rgb]{0.2,0.2,0.2}{11.58}          & \textcolor[rgb]{0.2,0.2,0.2}{13.55}                \\
\textcolor[rgb]{0.2,0.2,0.2}{OSS-ave}                    & \textcolor[rgb]{0.2,0.2,0.2}{\textbf{17.94}} & \textcolor[rgb]{0.2,0.2,0.2}{\textbf{22.47}}             & \textcolor[rgb]{0.2,0.2,0.2}{16.51}          & \textcolor[rgb]{0.2,0.2,0.2}{\textbf{21.25}}       \\ 
\midrule
                                                         & \multicolumn{2}{c|}{\textcolor[rgb]{0.2,0.2,0.2}{FFHQ-64~\cite{ffhq}}}                                              & \multicolumn{2}{c}{\textcolor[rgb]{0.2,0.2,0.2}{MSCOCO-512}~\cite{coco}}        \\ 
\hline
\textcolor[rgb]{0.2,0.2,0.2}{EDM}                        & \textcolor[rgb]{0.2,0.2,0.2}{18.68}          & \textcolor[rgb]{0.2,0.2,0.2}{23.24}                      & \textcolor[rgb]{0.2,0.2,0.2}{13.43}          & \textcolor[rgb]{0.2,0.2,0.2}{15.45}                \\
\textcolor[rgb]{0.2,0.2,0.2}{DDIM}                       & \textcolor[rgb]{0.2,0.2,0.2}{18.37}          & \textcolor[rgb]{0.2,0.2,0.2}{23.57}                      & \textcolor[rgb]{0.2,0.2,0.2}{12.31}          & \textcolor[rgb]{0.2,0.2,0.2}{14.58}                \\
\textcolor[rgb]{0.2,0.2,0.2}{FM}                         & \textcolor[rgb]{0.2,0.2,0.2}{17.72}          & \textcolor[rgb]{0.2,0.2,0.2}{21.31}                      & \textcolor[rgb]{0.2,0.2,0.2}{11.90}          & \textcolor[rgb]{0.2,0.2,0.2}{13.51}                \\
\textcolor[rgb]{0.2,0.2,0.2}{DM}                         & \textcolor[rgb]{0.2,0.2,0.2}{19.79}          & \textcolor[rgb]{0.2,0.2,0.2}{22.69}                      & \textcolor[rgb]{0.2,0.2,0.2}{13.50}          & \textcolor[rgb]{0.2,0.2,0.2}{14.73}                \\
\textcolor[rgb]{0.2,0.2,0.2}{GITS}                       & \textcolor[rgb]{0.2,0.2,0.2}{20.46}          & \textcolor[rgb]{0.2,0.2,0.2}{25.20}                      & \textcolor[rgb]{0.2,0.2,0.2}{10.60}          & \textcolor[rgb]{0.2,0.2,0.2}{11.07}                \\
\textcolor[rgb]{0.2,0.2,0.2}{OSS-ave}                    & \textcolor[rgb]{0.2,0.2,0.2}{\textbf{20.56}} & \textcolor[rgb]{0.2,0.2,0.2}{\textbf{25.21}}             & \textcolor[rgb]{0.2,0.2,0.2}{\textbf{14.10}} & \textcolor[rgb]{0.2,0.2,0.2}{\textbf{16.45}}       \\
\bottomrule
\end{tabular}
\vspace{-0.3cm}
\caption{PSNR results of different stepsize schedules.}
\label{table:compare}
\end{table}

    \begin{table*}[htbp]

    \centering
    \begin{tabular}{ccc|cccccc} 
    \toprule
    \textcolor[rgb]{0.2,0.2,0.2}{Method}   & steps & \textcolor[rgb]{0.2,0.2,0.2}{Overall}       & \begin{tabular}[c]{@{}c@{}}\textcolor[rgb]{0.2,0.2,0.2}{Single}~\\\textcolor[rgb]{0.2,0.2,0.2}{object}\end{tabular} & \begin{tabular}[c]{@{}c@{}}\textcolor[rgb]{0.2,0.2,0.2}{Two}~\\\textcolor[rgb]{0.2,0.2,0.2}{object}\end{tabular} & \textcolor[rgb]{0.2,0.2,0.2}{Counting}      & \textcolor[rgb]{0.2,0.2,0.2}{Colors}        & \textcolor[rgb]{0.2,0.2,0.2}{Position}      & \begin{tabular}[c]{@{}c@{}}\textcolor[rgb]{0.2,0.2,0.2}{Color }\\\textcolor[rgb]{0.2,0.2,0.2}{attribution}\end{tabular}  \\ 
    \hline
    \textcolor[rgb]{0.2,0.2,0.2}{Flow matching} & 100 & \textcolor[rgb]{0.2,0.2,0.2}{0.649}          & \textcolor[rgb]{0.2,0.2,0.2}{0.989}                                                                                  & \textcolor[rgb]{0.2,0.2,0.2}{0.803}                                                                               & \textcolor[rgb]{0.2,0.2,0.2}{0.722}          & \textcolor[rgb]{0.2,0.2,0.2}{0.774}          & \textcolor[rgb]{0.2,0.2,0.2}{0.220}          & \textcolor[rgb]{0.2,0.2,0.2}{0.390}                                                                                       \\ 
    \hline
    \textcolor[rgb]{0.2,0.2,0.2}{Flow matching} & 10  & \textcolor[rgb]{0.2,0.2,0.2}{0.590}          & \textcolor[rgb]{0.2,0.2,0.2}{0.963}                                                                                  & \textcolor[rgb]{0.2,0.2,0.2}{0.727}                                                                               & \textcolor[rgb]{0.2,0.2,0.2}{0.663}          & \textcolor[rgb]{0.2,0.2,0.2}{0.668}          & \textcolor[rgb]{0.2,0.2,0.2}{0.213} & \textcolor[rgb]{0.2,0.2,0.2}{0.310}                                                                                       \\
    \textcolor[rgb]{0.2,0.2,0.2}{DDIM} & 10  & \textcolor[rgb]{0.2,0.2,0.2}{0.623}          & \textcolor[rgb]{0.2,0.2,0.2}{0.971}                                                                                  & \textcolor[rgb]{0.2,0.2,0.2}{0.742}                                                                               & \textcolor[rgb]{0.2,0.2,0.2}{0.694}          & \textcolor[rgb]{0.2,0.2,0.2}{0.758}          & \textcolor[rgb]{0.2,0.2,0.2}{\textbf{0.22}} & \textcolor[rgb]{0.2,0.2,0.2}{0.353}                                                                                       \\
    \textcolor[rgb]{0.2,0.2,0.2}{GITS}     & 10  & \textcolor[rgb]{0.2,0.2,0.2}{0.604}          & \textcolor[rgb]{0.2,0.2,0.2}{\textbf{0.981}}                                                                         & \textcolor[rgb]{0.2,0.2,0.2}{0.732}                                                                               & \textcolor[rgb]{0.2,0.2,0.2}{0.678}          & \textcolor[rgb]{0.2,0.2,0.2}{0.697}          & \textcolor[rgb]{0.2,0.2,0.2}{0.210} & \textcolor[rgb]{0.2,0.2,0.2}{0.325}                                                                                       \\
    \textcolor[rgb]{0.2,0.2,0.2}{DM}     & 10  & \textcolor[rgb]{0.2,0.2,0.2}{0.643}          & \textcolor[rgb]{0.2,0.2,0.2}{0.971}                                                                         & \textcolor[rgb]{0.2,0.2,0.2}{\textbf{0.788}}                                                                               & \textcolor[rgb]{0.2,0.2,0.2}{0.719}          & \textcolor[rgb]{0.2,0.2,0.2}{\textbf{0.777}}          & \textcolor[rgb]{0.2,0.2,0.2}{0.188} & \textcolor[rgb]{0.2,0.2,0.2}{\textbf{0.415}}                                                                                       \\
    
    \textcolor[rgb]{0.2,0.2,0.2}{OSS-ave}     & 10  & \textcolor[rgb]{0.2,0.2,0.2}{\textbf{0.645}} & \textcolor[rgb]{0.2,0.2,0.2}{\textbf{0.981}}                                                                         & \textcolor[rgb]{0.2,0.2,0.2}{0.775}                                                                     & \textcolor[rgb]{0.2,0.2,0.2}{\textbf{0.728}} & \textcolor[rgb]{0.2,0.2,0.2}{\textbf{0.777}} & \textcolor[rgb]{0.2,0.2,0.2}{0.195}          & \textcolor[rgb]{0.2,0.2,0.2}{\textbf{0.415}}                                                                              \\
    \bottomrule
    \end{tabular}
    \vspace{-0.2cm}
    \caption{Results of Geneval based on the Flux.1-dev model with different sampling schedules.}
    \label{table:geneval}
    \end{table*}

\subsubsection{Optimal Stepsize as balanced task partition}
Our optimal stepsize schedule not only enables directly accelerating diffusion models sampling without finetuning network, but also inherently reflects the varying difficulty levels of denoising tasks across different noise intensities. This observation suggests that the optimal stepsize schedule can be interpreted as a principled multitask partitioning strategy, which may better balance the subtask relationships compared to conventional uniform timestep division, thereby mitigating distortion on final outputs. To validate this, we conduct experiments on ImageNet generation with the DiT-XL/2~\cite{dit} model by splitting the entire denoising process into five consecutive stages, with each stage modeled by independent parameters. Inspired by~\cite{salimans2022progressive, balaji2022ediff}, all sub-models are initialized with the pre-trained teacher model to accelerate convergence. The results shown in Table \ref{table:distill} demonstrate that our optimal stepsize can be regarded as a good task division, better allocating the model capacity. The result can be further improved when combined with other techniques, such as progressive distillation~\cite{salimans2022progressive}.

    \subsection{Compare with other methods}
    We compare our method with other stepsize strategies including ad-hoc schedules like DDIM, EDM, and flow matching, as well as dynamic adjustment methods such as DM~\cite{dm} and GITS~\cite{gits}. \Cref{table:compare} demonstrates our experimental results across four distinct datasets. Our method achieves results closest to those of the teacher model under 5 and 10 step generation setting. Moreover, we also validate our results on the GenEval~\cite{ghosh2023geneval} benchmark using Flux.1-dev~\cite{flux} as the foundation model. We list the quantitative results in Table ~\ref{table:geneval}, where the OSS-ave setting adopts 32 images for each category. We find that the optimal stepsize schedule achieves 10$\times$ acceleration compared to the 100-step teacher model with almost no performance degradation on the GenEval benchmark. As shown in Figure \ref{fig:geneval}, our method demonstrates the closest proximity to the teacher model comparing with other approaches.

    \section{Conclusion}
This paper proposes a dynamic programming framework for stepsize optimization in diffusion sampling. By reformulating stepsize scheduling as recursive approximation, our method derives theoretically optimal step sequences with low computational overhead. The experiments demonstrate universal robustness across architectures and solvers while maintaining output fidelity. Its seamless integration with existing direction optimization strategies enables practical deployment advantages. This work establishes an alternative pathway for efficient diffusion sampling.

\clearpage

{
    \small
    \bibliographystyle{ieeenat_fullname}
    \bibliography{main}
}

\clearpage
\setcounter{section}{0}
\renewcommand{\thesection}{\Alph{section}}

\section*{Appendix}\label{sec:appendix}
\addcontentsline{toc}{section}{Appendix}

    \section{Proof of the optimal substructure}\label{sec:optimal}
        Here, we aim to proof: 
        \emph{To solve the optimal approximation $\bm{z}[m][n]$, which indicate the $m$-step student outcome to the $n$-step teacher outcome, it always derives from $\bm{z}[m-1][k]$, where k $>$ n.}

    \begin{assumption}
        \label{assump} If $\bm{x}$ is closer to the teacher denosing result at $\bm{x}[i]$ than $\bm{x'}$, as the form
        $\norm{\bm{x} - \bm{x}[i]}^{2}\leq \norm{\bm{x'} - \bm{x}[i]}
        ^{2}$, we have:
        \begin{equation}
            \norm{\bm{v_{\theta,k}} - \bm{v_{\theta^*}}}^{2}_2\leq \norm{\bm{v_{\theta,k}'} - \bm{v_{\theta^*}}}
            ^{2}_2
        \end{equation}
    \end{assumption}

    We prove our theorem by the method of contradiction. Suppose there exist $\bm{x'}$ which is suboptimal than $\bm{z}[m-1][k]$ and satisfies:

    \begin{equation}
        \norm{\bm{x'} - \bm{x}[k]}^2_2 \ge \norm{\bm{z}[m-1][k] - \bm{x}[k]}^2_2
    \end{equation}
    But the optimal result at step $m$ is generated from the suboptimal result at timestep n, which satisfies:
    \begin{align} \label{Eq:target}
        & \norm{\mathcal{F}(\bm{x'}, \bm{v_\theta},\{k,n\}) - \bm{x}[n]}^2_2 \\
        & \leq \norm{\mathcal{F}(\bm{z}[m-1][k], \bm{v_\theta}, \{k,n\}) - \bm{x}[n]}^2_2 
    \end{align}
    We use $\bm{v'_{\theta}}$ represents the velocity of $\bm{x'}$ at timestep k, use  $\bm{v_{\theta}}$ represents the velocity of $\bm{z}[m-1][k]$ at timestep k:
    \begin{equation}
        \mathcal{F}(\bm{x'}, \bm{v_\theta},\{k,n\}) = \bm{x'} + \bm{v'_{\theta}}(n-k)
    \end{equation}
    Also, we assume the teacher solver results are an excellent ground truth approximation. We use $\bm{v_{\theta^*}}$ represents the ground truth velocity, which follows:
    \begin{equation}
        \bm{x}[n] = \bm{x}[k] + \bm{v_{\theta^*}}(n-k)
    \end{equation}

\begin{proof} 
    Here, we prove this in two cases. When the $\bm{x'}$ and $\bm{x}$ are closed to the teacher $\bm{x}[k]$, we have:

        \begin{align}
             & \norm{\mathcal{F}(\bm{x'}, \bm{v_\theta},\{k,n\}) - \bm{x}[n]}^2_2 \\
             & = \norm{\bm{x'}-\bm{x}[k] + (n-k)(\bm{v'_{\theta}} - \bm{v_{\theta^*}})}^2_2 \\
             & \approx \norm{(n-k)(\bm{v'_{\theta}} - \bm{v_{\theta^*}})}^2_2 \\
             & \geq   \norm{(n-k)(\bm{v_{\theta}} - \bm{v_{\theta^*}})}^{2}_2 \\
             & \approx \norm{\mathcal{F}(\bm{z}[m-1][k], \bm{v_\theta}, \{k,n\}) - \bm{x}[n]}^2_2 \\
        \end{align}
        This is contradictory to the hypothesis \cref{Eq:target}.

        When the $\bm{x'}$ and $\bm{x}$ are far from the teacher $\bm{x}[k]$, the input of the denosing network is far away from the training input. Therefore, we consider vectors $\bm{x'}-\bm{x}[k]$ and $\bm{v'_{\theta}} - \bm{v_{\theta^*}}$ to be statistically independent and treat them as random vectors in a high-dimensional space, thereby establishing their orthogonality. Then:
        \begin{align}
             & \norm{\mathcal{F}(\bm{x'}, \bm{v_\theta},\{k,n\}) - \bm{x}[n]}^2_2 \\
             & = \norm{\bm{x'}-\bm{x}[k] + (n-k)(\bm{v'_{\theta}} - \bm{v_{\theta^*}})}^2_2 \\
             & = \norm{\bm{x'}-\bm{x}[k]}^2_2 + \norm{(n-k)(\bm{v'_{\theta}} - \bm{v_{\theta^*}})}^2_2 \\ 
             & + 2*\langle \bm{x'} - \bm{x}[k], (n-k)(\bm{v'_{\theta}} - \bm{v_{\theta^*}}) \rangle \\
             & \approx \norm{\bm{x'}-\bm{x}[k]}^2_2 + \norm{(n-k)(\bm{v'_{\theta}} - \bm{v_{\theta^*}})}^2_2 \label{eq:inner_product} \\
             & \geq \norm{\bm{z}[m-1][k] - \bm{x}[k]}^2_2 +  \norm{(n-k)(\bm{v_{\theta}} - \bm{v_{\theta^*}})}^{2}_2 \\
             & \approx \norm{\mathcal{F}(\bm{z}[m-1][k], \bm{v_\theta}, \{k,n\}) - \bm{x}[n]}^2_2 \\
        \end{align}

        This is contradictory to the hypothesis \cref{Eq:target}. Therefore, the original hypothesis is not valid, the optimal results always come from the previous optimal results, which is the optimal substructure of our searching problems.
    \end{proof}

 \begin{algorithm}[t]
        \caption{Optimal stepsize for high order solver}
        \label{alg:algorithm2}
         \small
        \begin{algorithmic}
            [1]
            \Statex $\bm{D_\theta}$,$\bm{x}[N]$,$\bm{z}$ same as one order algorithm.
            \Statex { $od$: Inference Order}. 
            \Statex { $\bm{\mathcal{F}}:$ Forward function, $\bm{\mathcal{F}}$ $(x_{cur}, t_{cur}, \bm{v}, t_{next}, order)$}\\
             \textcolor{gray}{\# get teacher trajectory $\bm{x}[i]$ }
            \For{$\bm{i}$ = $N$ to 1} 
                \State{$v$ = $\bm{D_\theta}$$(\bm{x}[i],i)$}
                \State{ { $\bm{x}[i-1]$ = $\bm{\mathcal{F}}(\bm{x}[i], i, \bm{v}, i-1, od)$}}
            \EndFor \\
             \textcolor{gray}{\# use the student i steps result to approach the teacher's j steps result. }
            \For{$\bm{i}$ = 1 to $M$} 
                \State{$\bm{v}$ = $\bm{D_\theta}$$(\bm{z}[i-1,:],range(N,1))$}
                \State{$\bm{z_{tmp}}$ = ones$(od+1,N+ 1, N)$$\times$Inf}
                    \For{{$\bm{o}$ in $od$ to 1}} 
                        \For{$\bm{j}$ in $N$ to 1}
                            \State{$t_{next}$ = $range(0,j-1)$} 
                            \State{ $\bm{z_{tmp}}[o,j,:]$ = $\bm{\mathcal{F}}( \bm{z}[i-1,j], j, \bm{v}[j],t_{next},o)$}
                        \EndFor
                    \EndFor 
                \For{$\bm{j}$ in $N- 1$ to 0} 
                    \State{$\bm{cost}$ = $\bm{MSE}$($\bm{z_{tmp}}[:,:,j], \bm{x}[j]$)}
                    \State{{$\bm{r}[i,j]$ = $argmin(\bm{cost}.flatten())$}}
                    \State{{$\bm{z}[k,i]$ = $\bm{z_{tmp}}[\bm{r}[i.j]$//$(N+1),\bm{r}[i,j]\%(N+1),j]$}}
                    \State{$\bm{r}[i,j]$ = $\bm{r}[i,j] \% (N+1)$}
                \EndFor
            \EndFor 
            \textcolor{gray}{\# retrieve the optimal trajectory}
            \State{$\bm{R}$ = [0]} 
            \For{$i$ in $M$ to 1} 
                \State{$j$ = $\bm{R}[-1]$}

                \State{$\bm{R}$.append($\bm{r}[i][j]$)} 
            \EndFor 
            \State {\bfseries return} $\bm{R}[1:]$
        \end{algorithmic}
    \end{algorithm}

 \section{Searching steps with high order solvers} \label{sec:high_order}
    Here, we introduce the algorithm of searching optimal steps for high order solvers. A key distinction lies in the solver's dynamic order selection capability during the search phase: at each optimization step, the student solver may execute operations corresponding to the specified denoising order or any lower-order approximations. The optimal configuration is selected through minimum-distance alignment with the teacher trajectory across all denoising orders. After searching, the student solver utilize the optimal steps with the order given before. The complete optimization procedure is formalized in \cref{alg:algorithm2}.

 \section{Unification of different sampling trajectories}\label{sec:unify}
    This section elaborates on adapting various pre-trained models for sampling via flow matching. This process comprises three steps: transformation of network prediction targets, sampling trajectory alignment, and model input alignment. 
    The network prediction targets, comprising $\epsilon$, $x_0$, $v$, exhibit mutual convertibility through reparameterization. Our framework unifies their conversion to velocity $v$ through systematic transformation.
    
    Following the methodology in \cite{improverectifiedflow}, temporal alignment of sampling trajectories requires matching the signal-to-noise ratio (SNR) at each step. The flow matching diffusion process follows:
    \begin{equation}
        x_t = t\epsilon+(1-t)x_0
    \end{equation}
    while we assume the conventional forward trajectory of the pre-trained model adheres to:
    \begin{equation}
        x_t = \alpha_tx_0+\beta_t\epsilon
    \end{equation}
    Temporal alignment between flow matching time $t$ and reference trajectory time $t'$ is established through the ratio constraint:
  
    \begin{equation}
        \frac{1-t}{t} = \frac{\alpha_{t'}}{\beta_{t'}}
    \end{equation}
    Furthermore, the network input alignment between the current trajectory position $x_t$ and the pre-trained model's expected input $x_{t'}$ is achieved via linear transformation:
    \begin{equation}
        x_{t'} = \frac{\alpha_{t'}}{1-t}x_t
    \end{equation}
    These aligned coordinates ($t'$ and $x_{t'}$) enable direct utilization of pre-trained diffusion models to estimate the current trajectory velocity at timestep $t$, thereby enables sampling through flow matching.

\section{Visualization of optimal sampling schedule}\label{sec:oss}
\begin{figure}[t]
        \centering
        
        \includegraphics[width=0.5\textwidth]{
            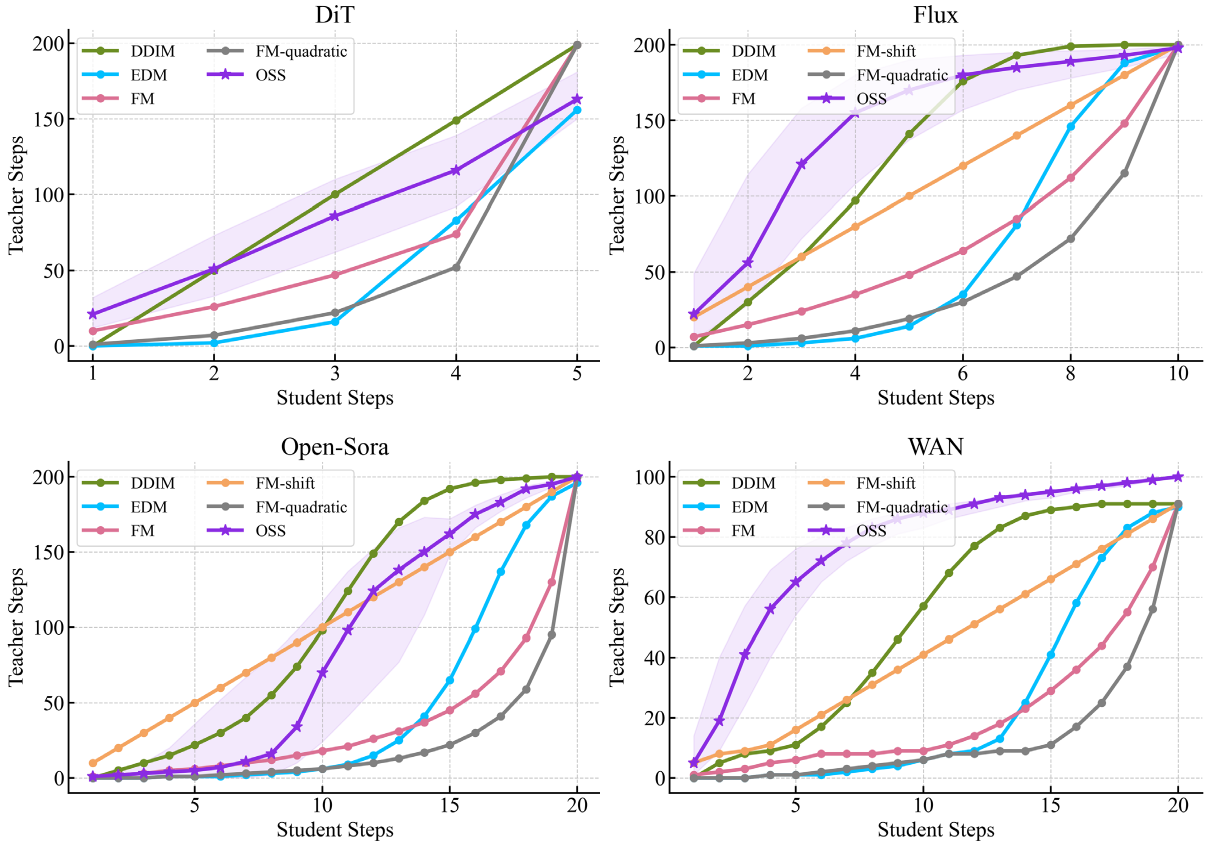
        }
        \caption{Different sampling schedules of DiT, Flux, Open-Sora, and WAN.} 
        \label{fig:stepschedule}
    \end{figure}

In this section, we present additional visualizations of sampling schedules. We maintain the original teacher schedule used in each open-source implementation: DiT~\cite{dit} employs DDIM~\cite{ddim} schedule, while Open-Sora~\cite{opensora}, Flux~\cite{flux}, and Wan~\cite{wan2025} adopt flow matching with timeshift~\cite{sd3} as their default schedules. We compare different few-step sampling strategies and utilize the unified framework in \Cref{sec:unify} to align timesteps with the teacher model. We present the teacher model's sampling timesteps corresponding to each few-step sampling strategy in \Cref{fig:stepschedule}.

    \section{More results} \label{sec:result}
    In this section, we provide more Flux.1-dev \cite{flux} generation results using 10 steps. Moreover, we also provide the video generation results in the github page.

    \begin{figure*}[t]
        \centering
        
        \includegraphics[width=0.9\textwidth]{
            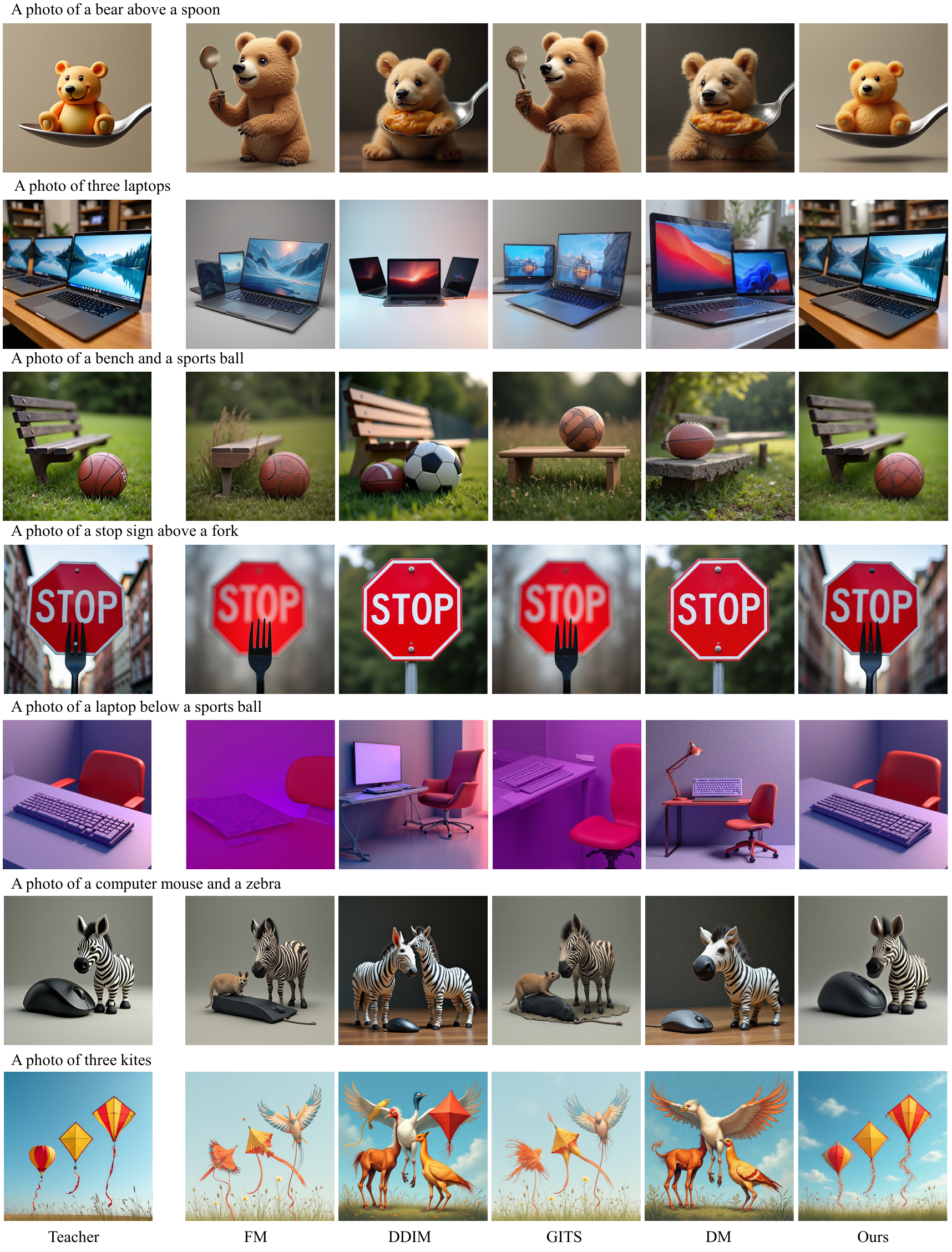
        }
        \caption{Flux generation results on Geneval.} 
 
    \end{figure*}

            \begin{figure*}[t]
        \centering

        \includegraphics[width=0.9\textwidth]{
            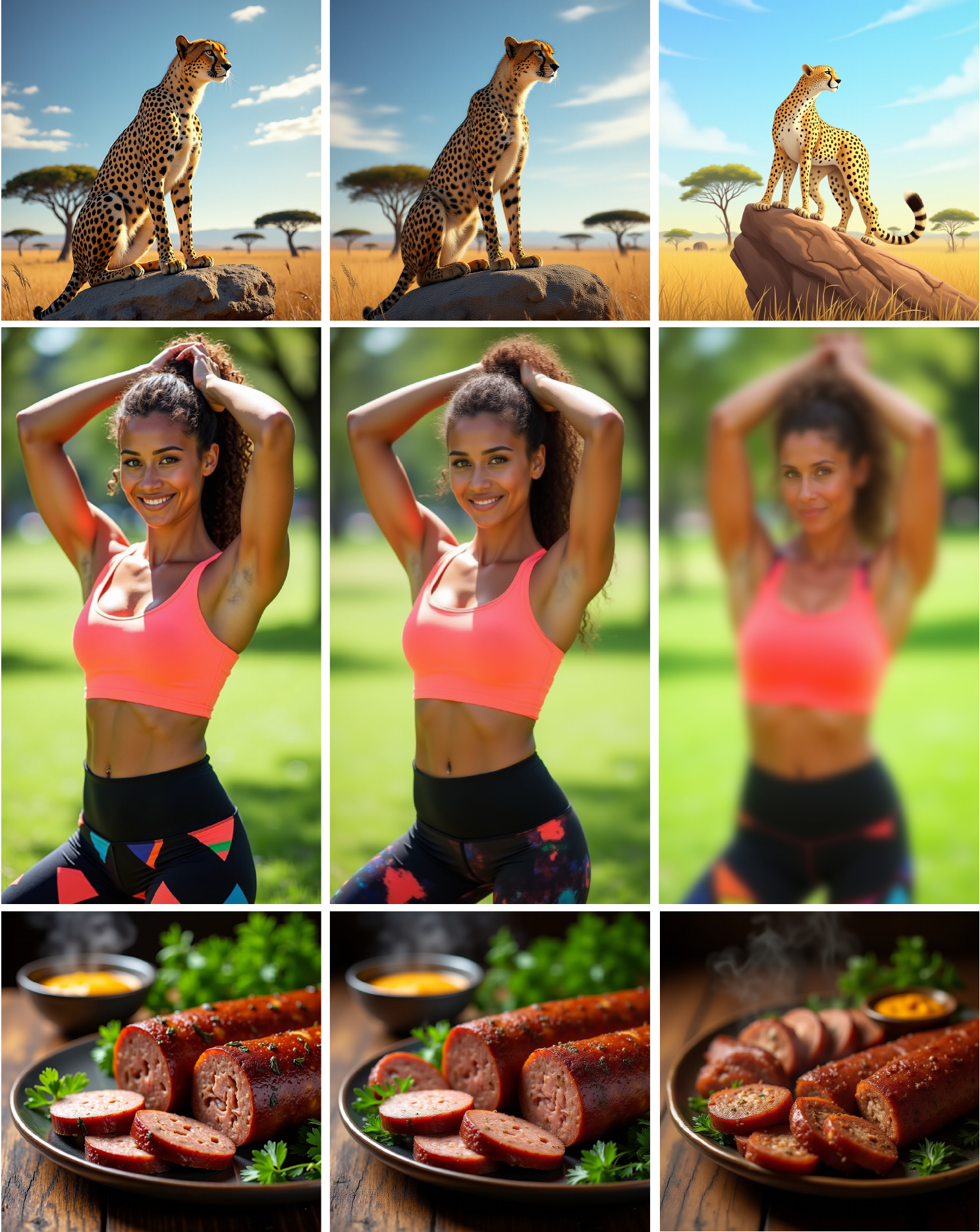
        }
        \caption{Flux generation results. Left: Original sampling result using 200 steps. Middle: Optimal stepsize sampling result within 10 steps. Right: Naively reducing sampling steps to 10.} 
 
    \end{figure*}

         \begin{figure*}[t]
        \centering

        \includegraphics[width=0.9\textwidth]{
            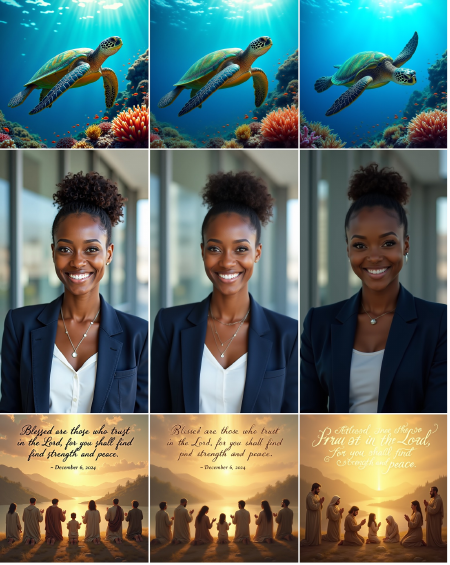
        }
        \caption{Flux generation results. Left: Original sampling result using 200 steps. Middle: Optimal stepsize sampling result within 10 steps. Right: Naively reducing sampling steps to 10.} 

    \end{figure*}
    
         \begin{figure*}[t]
        \centering
\includegraphics[width=0.9\textwidth]{
            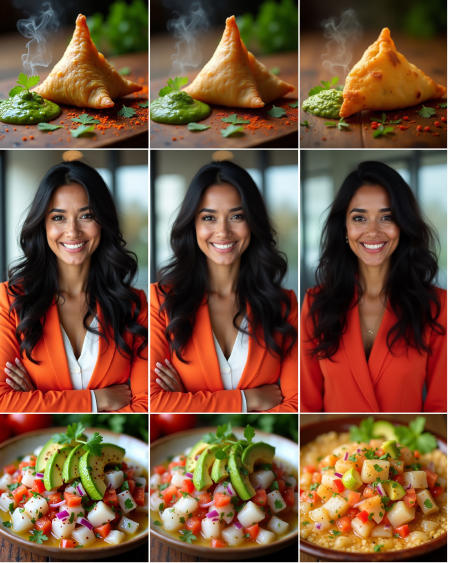
        }
        \caption{Flux generation results. Left: Original sampling result using 200 steps. Middle: Optimal stepsize sampling result within 10 steps. Right: Naively reducing sampling steps to 10.} 
 
    \end{figure*}

\end{document}